\documentclass[runningheads]{llncs}

\usepackage[utf8]{inputenc} 
\usepackage[T1]{fontenc}    
\usepackage{lmodern}
\usepackage{hyperref}       
\usepackage{url}            
\usepackage{booktabs}       
\usepackage{nicefrac}       
\usepackage{microtype}      

\usepackage{latexsym,amssymb,amstext,amsfonts,amsmath,setspace,mathtools,bm}
\usepackage{subcaption}
\usepackage{enumerate} 
\usepackage{paralist}
\usepackage{float}
\usepackage{array}
\usepackage{comment}

\usepackage{tikz}
\usetikzlibrary{fit,positioning}
\usetikzlibrary{calc}

\definecolor{nice_blue}{RGB}{65, 105, 225}
\definecolor{nice_red}{RGB}{168, 34, 34}
\definecolor{IGCBlue}{HTML}{16197A}
\definecolor{dark_green}{RGB}{20, 110, 10}

\definecolor{graph_blue}{RGB}{144, 195, 212}
\definecolor{graph_purple}{RGB}{195, 144, 212}
\definecolor{graph_green}{RGB}{161, 212, 144}
\definecolor{graph_orred}{RGB}{212, 161, 144}

\DeclareMathOperator*{\argmin}{arg\,min}

\DeclareMathOperator*{\diag}{diag}
\DeclareMathOperator*{\tr}{tr}

\usepackage{graphicx}
\begin{document}
\title{Equivariant Filters for Efficient Tracking in {3D} Imaging}
%
%
\author{Daniel Moyer$^{1}$, Esra Abaci Turk$^{2}$, P Ellen Grant$^{2}$, William M. Wells$^{1,3}$, and Polina Golland$^{1}$}
%
\authorrunning{D. Moyer et al.}
%
\institute{
$^{1}$CSAIL, Massachusetts Institute of Technology, Cambridge, MA, USA\\
$^{2}$Boston Children’s Hospital, Harvard Medical School, Boston, MA, USA \\
$^{3}$Brigham and Women’s Hospital, Harvard Medical School, Boston, MA, USA
}
%
\maketitle              
\begin{abstract}

We demonstrate an object tracking method for {3D} images with fixed computational cost and state-of-the-art performance. Previous methods predicted transformation parameters from convolutional layers. We instead propose an architecture that neither flattens convolutional features nor uses fully connected layers, but instead relies on equivariant filters to preserve transformations between inputs and outputs (e.g., rotations/translations of inputs rotate/translate outputs). The transformation is then derived in closed form from the outputs of the filters. This method is useful for applications requiring low latency, such as real-time tracking. We demonstrate our model on synthetically augmented adult brain MRI, as well as fetal brain MRI, which is the intended use-case.

\end{abstract}
\section{Introduction}

Real-time prospective tracking and slice prescription in volumetric modalities like MRI become possible as registration speeds approach slice acquisition/safety-prescribed cooldown times \cite{white2010promo,malamateniou2013motion}. If rigid landmarks are easily identifiable, rigid tracking in volumes reduces to automatic tagging or segmentation. However, in low signal-to-noise (SNR), low resolution applications such as fetal imaging, standard anatomic landmarks may not be identifiable from images; it is helpful in those cases to learn features that can be robustly detected. Moreover, if utility is directly proportional to speed (e.g., for navigator scans),
SNR and resolution are reduced in favor of lower scan times.
In this regime anatomic intuition may no longer match imaging conditions, and relevant features may be more robust if learned from images.

Recent learning based methods for rigid and affine registration have used convolutional neural networks (CNNs), which generally have faster inference-time speed than traditional image-grid iterative algorithms.
CNN-based registrations almost universally regress transformation parameters directly by applying a set of stacked convolutional
filters to images followed by a series of fully connected layers.
Such architectures are common in other vision tasks \cite{bengio2017deep}.

Convolutions are translation equivariant: as inputs shift, output features shift accordingly. However, the fully connected network that estimates transformation parameters must learn from data a similar property between shifts and output estimates; the network structure does not naturally give rise to these symmetries. Moreover, neither structure encodes these properties naturally for rotation, and thus the equivariances must be learned. This is inefficient in data and computation, and, as we show, induces sub-optimal performance.


Instead, we constrain the feature extraction portion of networks (i.e., the convolutional filters) to be both translation and rotation equivariant. If the activation maps shift and rotate with the target object, we can avoid the fully connected regression step and instead compute least squares optimal transformations in closed form from summary statistics of the activation maps.
We introduce a novel architecture for rigid tracking that leverages these two techniques. Our proposed method has the same computational costs as regular convolution at inference time, and we show that it provides more accurate tracking on human adult brain MRI and our intended use-case of fetal brain MRI.

\subsection{Previous Work}

Real-time object tracking (rigid registration) in volumetric imaging has a long history \cite{cox1999real,woods1998automated}.
Early real-time methods in MRI and PET relied on either voxelwise intensity matching and gradient descent on parameters \cite{cox1999real,thesen2000prospective}, or intrinsic properties of the image acquisition in k-space e.g., the oversampling of low frequencies in PROPELLER \cite{pipe1999motion}.
For adult human heads constrained by the physical geometry of the head coil, observed shifts and rotations are small in scale (e.g., maximally 10 degrees \cite{thesen2000prospective}), and assumed to be global shifts of the entire non-zero acquisition \cite{pipe1999motion}.
Later systems for adult human heads include ``navigator'' scans and Kalman filters\cite{white2010promo}, but still consider a range of motion reasonable for scanning adults.

In fetal imaging, classical tracking methods such as PROPELLER fail in practice \cite{malamateniou2013motion}. Iterative intensity matching is feasible but tends to be slower due to the extended search space \cite{chee2018airnet} and multiple scales of image gridding. 

Current methods in rigid and affine alignment are built almost exclusively on convolutional networks \cite{chee2018airnet,sloan2018learning,de2019deep}. As they have a fixed number of operations which are amicable to GPU parallelization and compiler optimizations \cite{chen2018TVM}.
Salehi et al. 2018 \cite{salehi2018real} use a regression similar to image prediction architectures \cite{bengio2017deep}, where the reference and moving images are concatenated and provided as input to a commonly used architecture of alternating convolution and pooling. This leads to a flattening followed by a series of fully connected layers which estimate transformation parameters.
Chee et al. 2018 \cite{chee2018airnet} and de Vos et al. 2019 \cite{de2019deep} instead use convolution layer stacks with separate inputs but shared weights.

We avoid the previous paradigm of stacked convolution to direct regression; rigid motion has known analytic symmetries, properties which are ignored by the structure of fully connected layers, and must be relearned in their parameters. These same properties we exploit in our proposed method.

We emphasize that our work is unrelated to the slice-to-volume registration problem~\cite{ferrante2017slice}, as our goal is to align volumetric 3D images.

\begin{figure}[t!]
\begin{center}
\includegraphics[width=1.0\textwidth]{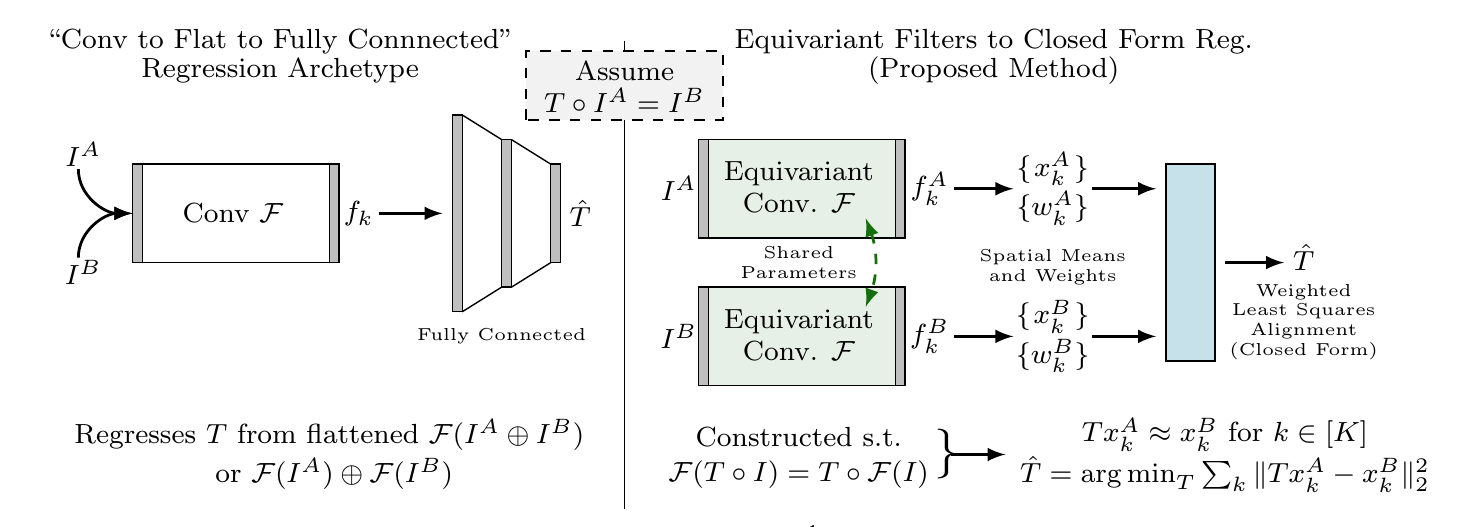}
\caption{Our network method, along with the convolution archetypes. }
\label{fig:net-diagram}
\end{center}
\end{figure}

\section{Method}


We assume that two given images $I^A$ and $I^B$ are related by an unknown rigid transformation T, i.e.,
\begin{align}
T \circ I^A = I^B.
\end{align}
A filter bank $\mathcal{F}:\mathcal{I} \rightarrow \mathcal{I}(\mathbb{R}^+)^K$ with $K$ non-negative real-valued channels is called equivariant\footnote{More general forms of equivariance are defined in the literature \cite{weiler20183d}, but this definition is sufficient for our purposes. We do not use properties of any other form.} under rigid transformations if for each channel $\mathcal{F}_k$ of $\mathcal{F}$,
\begin{align}
\mathcal{F}_k( T \circ I ) = T \circ \mathcal{F}_k(I).
\end{align}
In other words, $\mathcal{F}$ is equivariant if the outputs of $\mathcal{F}$ translate and rotate as the inputs are rotated and translated. Assuming $\mathcal{F}$ is equivariant and denoting $\mathcal{F}(I)=f$, we have
\begin{align}
f_k^B = \mathcal{F}_k (I^B) = \mathcal{F}_k ( T \circ I^A ) = T f_k^A. \label{eq:fBeqTfA}
\end{align}

Since filter outputs $\{f_k^A\}$ and $\{f_k^B\}$ are in correspondence, we expect specific summary statistics (e.g., spatial mean activation point) to similarly match across images. 
Extracting the statistics from the filter outputs yields point clouds $\{x_k^A\}$ and $\{x_k^B\}$ with correspondence. $T x_k^A  = x_k^B$ for $k \in \{1,\dots,K\}$ up to discretization and resampling error. Thus, $T$ can be estimated from $\{x_k^A\}$ and $\{x_k^B\}$ using for example the least squares formulation \cite{arun1987least,horn1987closed,kabsch1976solution}.

Our proposed method implements the above procedure. It applies equivariant filters to each volume, computes the spatial mean of each filter, and then
estimates the transformation that minimizes squared error, which in turn provides the $\ell_2$ optimal alignment of those spatial means between the volumes (Figure \ref{fig:net-diagram}).
The learned parameters remain entirely in the equivariant filter construction; after filter outputs are collected all subsequent computations are analytically prescribed. The entire structure is differentiable, thus a loss function on the outputs can be backpropagated in order to train the equivariant filter parameters. Losses may be defined either on transformations or on image differences under the transformations (i.e., $d(\hat{T}\circ I^A, I^B)$ for image metric $d$).  

\subsection{Construction of Equivariant Convolutional Filters}

Previously we defined $\mathcal{F}$ generally; however, $\mathcal{F}$ must be an efficient but expressive differentiable operation on image arrays. This leads us to stacked convolutional filters, which are already intrinsically equivariant to translation. In order to ensure rotational equivariance, we follow the construction in Weiler et al. \cite{weiler20183d}, which retains translation equivariance.

Any rigid transformation $T$ can be decomposed into translation $t$ and rotation $R$. Given a field $F(x)$ over the spatial domain, the transformation operator is $\rho(R)F(R^{-1}(x-t))$, where the function $\rho$ depends on the order of the field. While the images we consider and our desired outputs are both scalar fields with trivial $\rho$ functions, for more expressive equivariant filters we require higher order fields (e.g., vector or matrix valued fields). Conventional multi-channel image operations use all scalar fields.

Weiler et al. \cite{weiler20183d} showed that convolutions between fields (``cross-correlation'' in other settings) can be expressed as a block diagonal operation between specific basis components (irreducible representations), and that rotation equivariant kernels are a linear subspace of the possible kernels. The authors then construct a basis of such kernels analytically using spherical harmonics. Convolutions expressed in this basis are weighted combinations of the basis elements, which are analytically determined sub-blocks of conventional convolution kernels. These may be pre-computed for given field orders and discrete kernel widths. The weights between the basis elements form the learnable factors.


For a single channel (i.e., a scalar field), equivariant kernels must be isotropic (i.e., having spherical isoclines). However, grouping multiple channels together into higher order fields allows for non-isotropic equivariant kernels, acting on sets of channels. Higher order convolutional fields also require specialized non-linearities, but these are operationally the same as normal convolutions, and also without learned parameters. We refer the reader to \cite{weiler20183d,smidt2021finding} for details. We stack alternating layers of equivariant convolutions and their corresponding non-linearities to form $\mathcal{F}$, our equivariant filter bank.





\subsection{Registration of Equivaritant Filters}



We reduce each non-negative filter response $f_k$ to its mean spatial position
\begin{align}
x_k = \frac{1}{N_1 N_2 N_3}\sum_{v} v f_k(v)
\end{align}
for images with dimension $N_1\times N_2 \times N_3$, where $v_{\ell,m,n}$ is the spatial coordinate of image index $(\ell,m,n)$, and $f_k(v_{\ell,m,n})$ is the corresponding value for the $k^{th}$ filter. Since $f_k^B = T f_k^A$, the spatial means of the filters will also be related similarly: $x_k^B \approx Tx_k^A$. Here the error is due to discretization. This over-determined linear system is then solved for rigid transformation $T$, which we can express as the least squares problem:
\begin{align}
\hat{T} = \argmin_{T} \sum_k\| x_k^B - Tx_k^A\|_2^2 .
\end{align}
Analytic solutions have been described multiple times \cite{arun1987least,horn1987closed,kabsch1976solution}. We further introduce non-negative channel importance weights $w_{k}^A$ and $w_{k}^B$, in order to reduce the influence of less important filters; we use the total power of each channel, e.g., $\tilde{w}^A_k = \sum_{v} f_k^A(v)$ as proposal weights (filter outputs are non-negative), then normalize to form $w_k^A = \tilde{w}_k^A / \sum_{k'} w_k'^A$, and similarly for $\{w_k^B\}$. From these we form generic channel weights $w_k = w_k^A w_k^B$, which we add to our optimization:
\begin{align}
\hat{T} = \argmin_{T} \sum_k w_k \| x_k^B - Tx_k^A\|_2^2 .
\end{align}
As shown in \cite{horn1987closed}, this also has an analytic solution similar to the unweighted problem. Computing weighted centroids $c^A = \sum_k x_{k}^A w_k$ and $c^B = \sum_k x_{k}^B w_k$, the optimal translation is $\hat{t} = c^B - c^A$. 
Re-centering each point $\bar{x}_{k}^A = x_{k}^A - c^A$ and stacking the $\bar{x}_{k}^{A}$ into matrices $X_A$ and $X_B$, the optimal rotation is $\hat{R} = VU^T$, where $(\bar{X}_A)^TW\bar{X}_B = UDV^T$ is the SVD of the weighted cross correlation matrix and $W = \diag(w_1,\dots,w_k)$.




\subsection{Loss Functions and Implementations}

Our architectures prescribe a forward operation. If the transformation parameters are known, loss functions can be defined for the output rotation and translation parameters. For translation parameters, the $\ell_2$ loss is natural, but for rotation parameters more nuanced metric spaces can also be used. Salehi et al. 2018 \cite{salehi2018real} propose a geodesic loss on the rotation parameter matrix
\begin{align}
\mathcal{L}_{\mathrm{geo}} = \arccos\left\{ (\tr\tilde{R}R - 1)/2 \right\}.
\end{align}
Zhou et al. \cite{zhou2019continuity} propose a projection and renormalization of the first two rows of each rotation matrix (``6D-loss'')
\begin{align}
\mathcal{L}_{\mathrm{6D}} = \| ~( ~ \tilde{R}_{:,1:2} / \|\tilde{R}_{:,1:2} \|_2 ~) - ( ~ R_{:,1:2}/\|R_{:,1:2}\|_2 ~ )~\|_2
\end{align}
which they show to be continuous for all elements in $SO(3)$.

In the absence of known transformation parameters we may also use image losses under the estimated transformation, i.e.,
\begin{align}
\mathcal{L}_{\mathrm{Image}} = d( T \circ I^A, I^B ).
\end{align}
Given a loss choice, we back-propagate errors and fit parameters by standard gradient based optimization \cite{bengio2017deep}.


\section{Experiments}

\begin{figure}[t!]
\begin{center}
\includegraphics[width=0.9\textwidth]{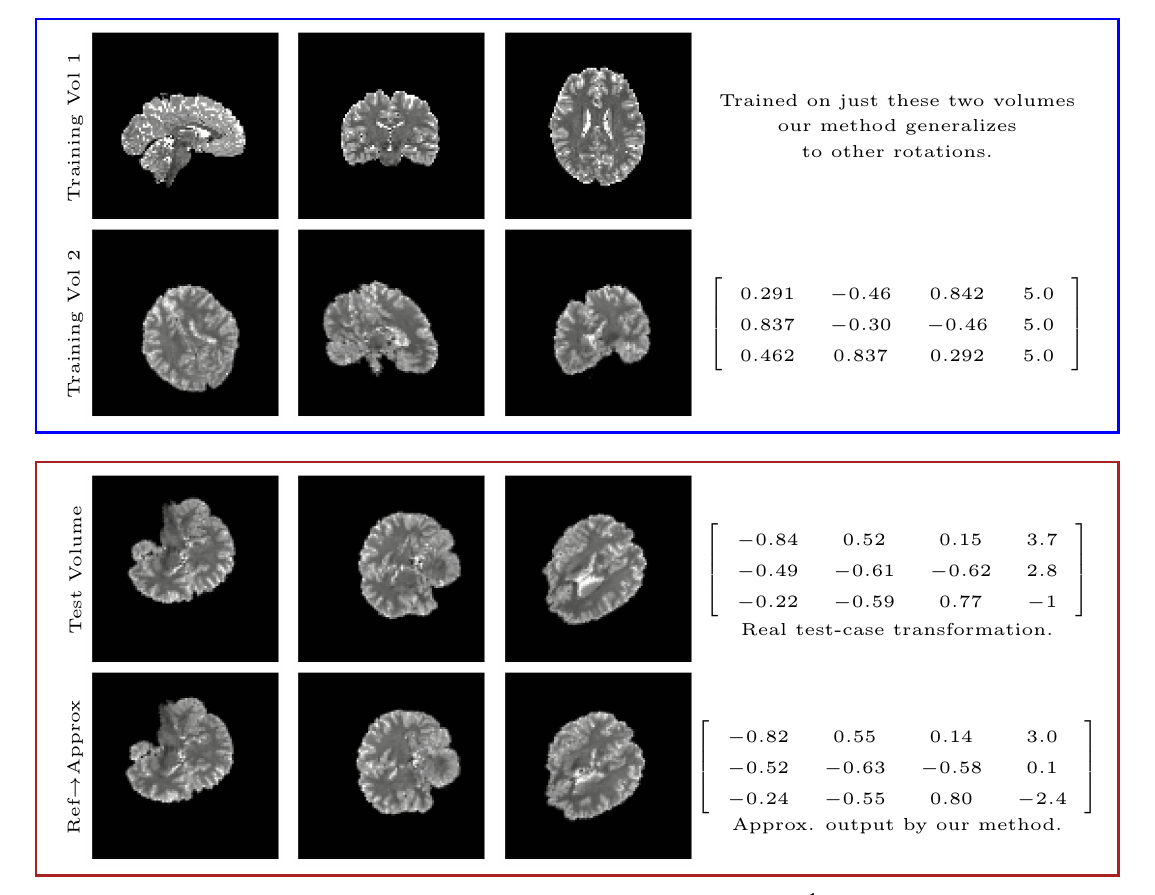}
\caption{The training and results for our first experiment. At top in the {\color{blue}blue} box we show central slices of the entire training dataset: a single subject in two poses. Below in the {\color{red}red} box we first show the central slices of the same subject in a novel pose, then in the second row we show a reference volume transformed by our estimated parameters. True and estimated parameters are shown at right.}
\label{fig:overfit}
\end{center}
\end{figure}

\begin{table}[t!]
\begin{center}
\begin{tabular}{| c | c | c | c | c |}
\hline
\multicolumn{2}{| c | }{ Angular Error } & \multicolumn{2}{  c | }{ Translational Error }  & Dice Index\\ 
\hline
Euler Angle & $\| I - R\hat{R}^T\|_F$ & mm & Voxels & -- \\ \hline
$2.0^\circ \pm 1.6$ & $0.07 \pm 0.03$ & $6.1 \pm 2.0$ & $2.2 \pm 0.7$ & $0.96 \pm 0.01$ \\ \hline
\end{tabular}
\vspace{0.5em}
\caption{Mean and std. of error measures for the first experiment averaged over 100 novel poses. The proposed method was trained on a single pair of images. The Euler Angle error is the Mean Abs. Error in degrees averaged over the axes.}
\label{tab:overfit}
\end{center}
\end{table}

\begin{figure}[h!]
\begin{center}
\includegraphics[width=0.9\textwidth]{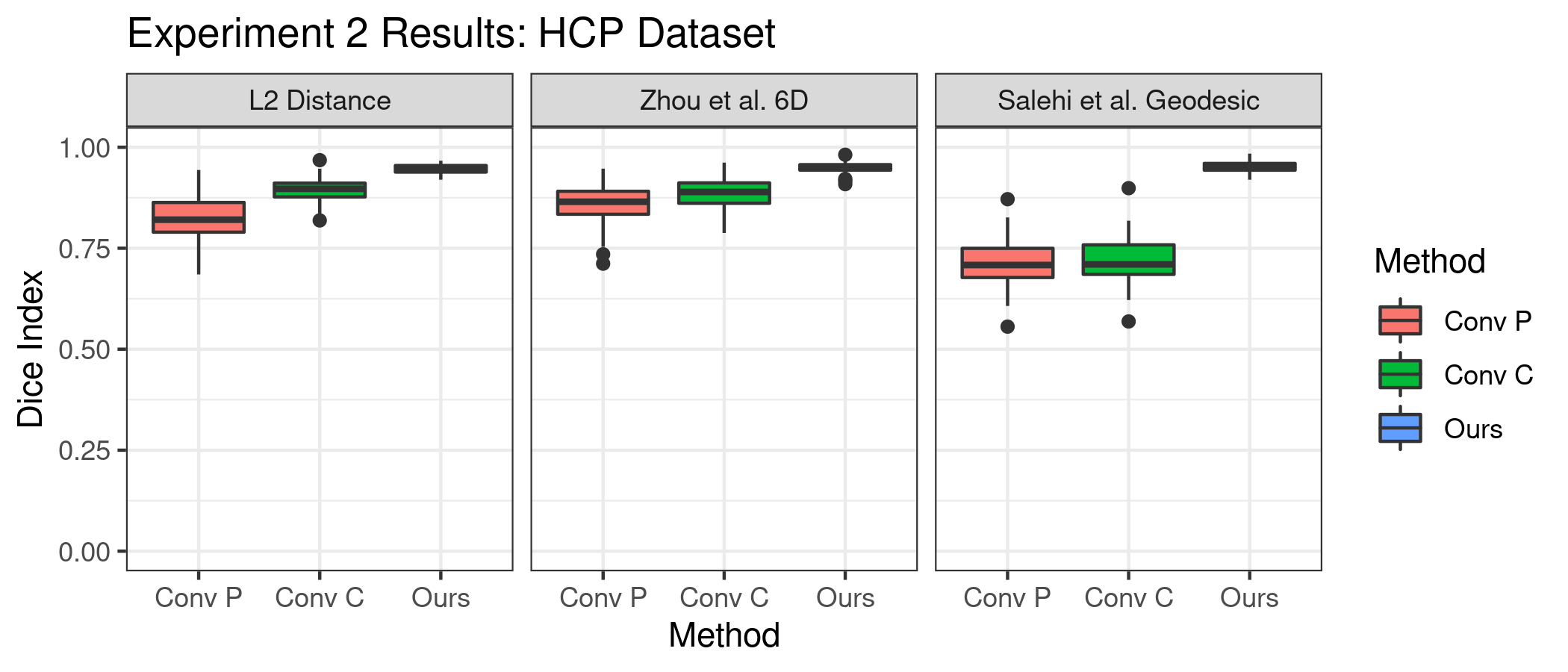}
\includegraphics[width=0.9\textwidth]{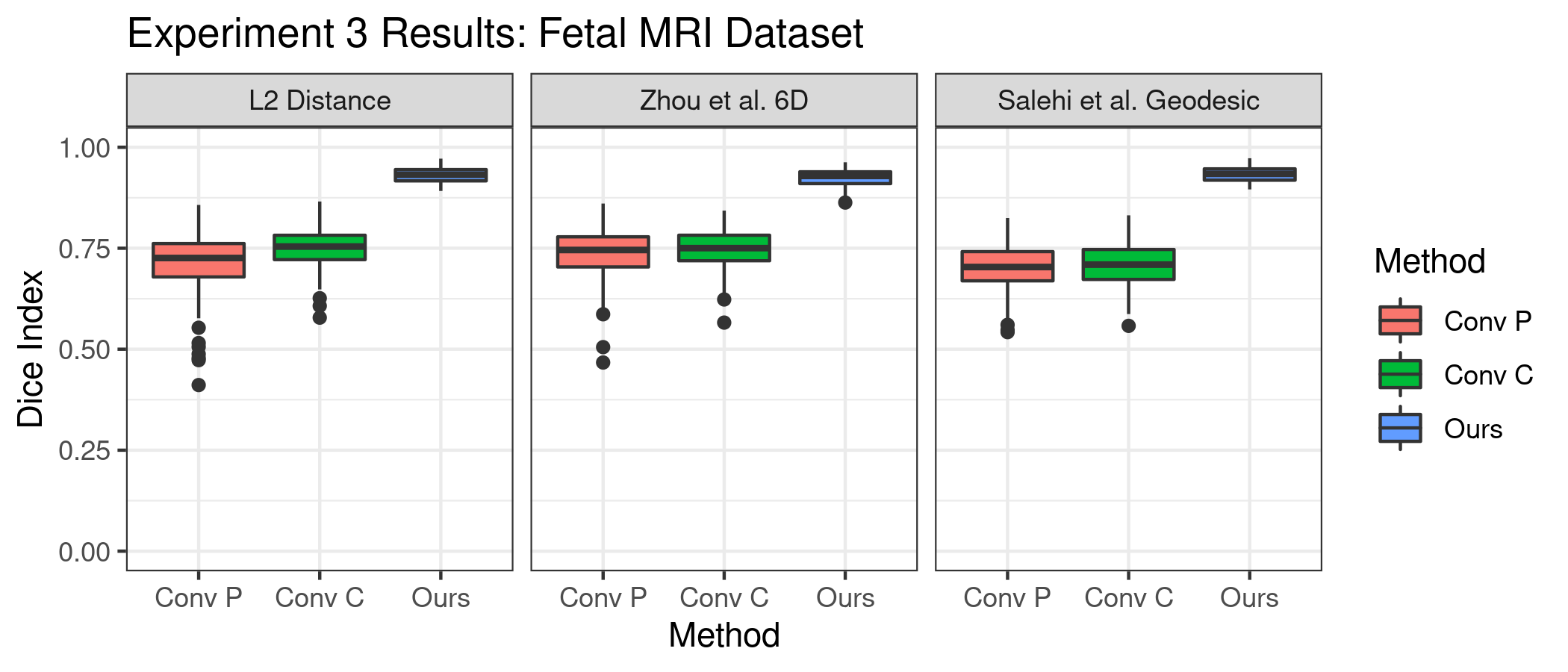}
\caption{Testing results for the second experiment on HCP data (\textbf{top row}) and for the third experiment with fetal MRI data (\textbf{bottom row}), measured by Dice Index overlap of masks under estimated transformations between images. Each inset indicates training under a different loss ($\ell_2$, $\mathcal{L}_{\mathrm{6D}}$ \cite{zhou2019continuity}, and $\mathcal{L}_\mathrm{geo}$ \cite{salehi2018real}), for all methods. \textbf{Higher} is better, with 1 indicating perfect overlap. \textbf{Conv P} is comparable to \cite{chee2018airnet,de2019deep}, while \textbf{Conv C} is comparable to \cite{salehi2018real}.}
\label{fig:results}
\end{center}
\end{figure}

We demonstrate our method in three separate empirical experiments, two on subsets of the Human Connectom Project (HCP) young adult cohort \cite{van2013wu}, and one on a fetal imaging dataset. The HCP dataset consists of subjects' T2-weighted brain volumes, downsampled from 0.7 mm isotropic to 2.8 mm isotropic, and then padded to a 96x96x96 voxel volume. Images were masked for brain tissue and histograms normalized by percentile filter; training-testing splits as well as data augmentation vary by experiment. The fetal dataset consists of MRI time-series from 53 healthy mothers pregnant with healthy singletons at gestational ages ranging from 25 to 35 weeks. MRI were acquired on a 3T Skyra Scanner (Siemens Healthcare, Erlangen, Germany). Multislice single-shot gradient echo EPI sequences were acquired at 3mm isotropic resolution, with a mean matrix size (field-of-view) of 120x120x80 which we crop to 64x64x64, TR=5-8s, TE=32-38ms, and 90 degree flip angles. Automatic brain masking was applied and dilated by 4 voxels to reflect uncertainty, after which images were intensity normalized.

In all three experiments training, validation, and testing datasets are completely disjoint, and in the latter two experiments subjects are not repeated in separate poses between datasets i.e., no individuals are shared between training/validation/testing sets. For both datasets and all three experiments we train both baseline methods and variations of our proposed method using the Adam optimizer \cite{kingma2014adam} for 2000 epochs, taking the best performing parameters and hyperparameters with respect to a validation set.  For all experiments batch size was set to 1. Three separate losses were tested: $\mathcal{L}_{\mathrm{geo}}$ \cite{salehi2018real}, $\mathcal{L}_{\text{6D}}$ \cite{zhou2019continuity}, and $\ell_2$.

\textbf{Parameters: }After several preliminary experiments we selected a stack of 5 equivariant convolutions, all with 5x5x5 kernels, using field-appropriate ReLU non-linearities throughout, and $16$, $16$, and $4$ fields of orders 0, 1, and 2 respectively, except for the final layer which has 64 scalar channels.

\textbf{Baselines:} We compare our method against two convolutional baseline architectures, one using a combined processing stream (``\textbf{Conv C}'') as found in \cite{salehi2018real}, and one using separate, parallel convolutional processing streams with tied weights (``\textbf{Conv P}''), as found in both \cite{chee2018airnet} and \cite{de2019deep}. For each method, we searched over the number of layers and channels per layer for optimal configurations given compute resources. We found the following to be optimal for our tests: 3 conv layers of 3x3x3 kernels with 64 channels each, interspersed with 2x2x2 pooling, followed by fully connected layers of 1024, 512, and 256 units, using ReLU activations throughout except for the output layer. The initial fully-connected layer has more parameters than the entirety of our network.


\textbf{Single Subject Experiment:}  In order to demonstrate the structural advantage of our method, we first train on \emph{just two volumes} of one subject. As Fig. \ref{fig:overfit}, and Table \ref{tab:overfit} report, from this single example of a rotation our method generalizes well to new poses of the same subject. 

\textbf{Group HCP Experiment:} Next we train the network on 500 subjects from the HCP dataset, holding out 100 for validation and 100 for testing, in order to test the generalization of different methods across subjects. As shown in the \textbf{top row} of Fig. \ref{fig:results}, our method performs better than both baselines, though all methods work reasonably well using either $\ell_{2}$ or $\mathcal{L}_{\mathrm{6D}}$ losses.

\textbf{Fetal Experiment:} Our third experiment trains the proposed method on the fetal MRI dataset, using 33 subjects for training (50 volumes per subject), 4 subjects held out for validation, and 16 subjects for testing (10 volumes per subject). As shown in the \textbf{bottom row} of Figure \ref{fig:results}, our method significantly outperforms the baseline methods for this dataset. This may be due to the smaller and more highly varying size of the fetal brains.


\section{Discussion and Conclusion}

We have described an equivariant filter based method for rigid tracking suitable for applications where speed is paramount. As we have shown empirically, our approach is demonstrably superior to methods based on conventional convolution. Moreover, due to the equivariant property of the filters, it generalizes to unseen poses by design. While a complete prospective motion correction system will have multiple other components (e.g., reconstruction, denoising, masking, slice proscription), tracking is a key step in this process, and accurate and efficient methods for tracking such as the one introduced here are thus of value for building such systems. 

\subsection*{Acknowledgements}

This work was supported by NIH NIBIB NAC P41EB015902, NIH NICHD R01HD100009, Wistron Corporation, and the MIT-IBM Watson AI Lab.


%
%
%
\bibliographystyle{splncs04}
\bibliography{eq_rigid_reg}
%




\end{document}